MULTI-AGENT SIMULATION AND MANAGEMENT PRACTICES

Peer-Olaf Siebers* and Uwe Aickelin

School of Computer Science & IT (ASAP)

University of Nottingham

Nottingham, NG8 1BB

UK

Email: pos@cs.nott.ac.uk; uxa@cs.nott.ac.uk

Helen Celia and Chris Clegg

Centre for Organisational Strategy, Learning & Change, Leeds University Business School

University of Leeds

Leeds, LS2 9JT

UK

Email: h.celia@leeds.ac.uk; c.w.clegg@leeds.ac.uk

MULTI-AGENT SIMULATION AND MANAGEMENT PRACTICES

INTRODUCTION

Intelligent agents offer a new and exciting way of understanding the world of work. Agent-Based Simulation (ABS), one way of using intelligent agents, carries great potential for progressing our understanding of management practices and how they link to retail performance. We have developed simulation models based on research by a multi-disciplinary team of economists, work psychologists and computer scientists. We will discuss our experiences of implementing these concepts working with a well-known retail department store.

There is no doubt that management practices are linked to the performance of an organisation (Reynolds et al., 2005; Wall & Wood, 2005). Best practices have been developed, but when it comes down to the actual application of these guidelines considerable ambiguity remains regarding their effectiveness within particular contexts (Siebers et al., forthcoming a).

Most Operational Research (OR) methods can only be used as analysis tools once management practices have been implemented. Often they are not very useful for giving answers to speculative 'what-if' questions, particularly when one is interested in the development of the system over time rather than just the state of the system at a certain point in time.

Simulation can be used to analyse the operation of dynamic and stochastic systems. ABS is particularly useful when complex interactions between system entities exist, such as autonomous decision making or negotiation. In an ABS model the researcher explicitly describes the decision process of simulated actors at the micro level. Structures emerge at the macro level as a result of the actions of the agents and their interactions with other agents and the environment.

We will show how ABS experiments can deal with testing and optimising management practices such as training, empowerment or teamwork. Hence, questions such as "will staff setting their own break times improve performance?" can be investigated.

BACKGROUND

Our current work examines the UK retail sector, but what we are learning about system modelling will be useful for modelling any complex system that involves many human interactions and where the actors work with some degree of autonomy.

The effects of socially embedded constructs, such as management practices, are inherently linked to their context. For many years social scientists have struggled to delineate the effect of management practices in order to reliably draw linkages to performance measures, and they continue to do so (e.g. Wall & Wood, 2005). The application of novel OR methods is necessary to reveal system-level effects of the introduction of specific management practices. This holds particularly when the focal interest is the development of the system over time, as in the real world.

Existing Tools for Modelling the Impact of Management Practices

As a result of a literature review we have undertaken (Siebers et al., forthcoming b) we found that only limited work has been conducted on developing models that allow investigation of the impact of management practices on retail performance. Most papers that investigate retail performance focus primarily on consumer behaviour and efficiency evaluation of firms with less emphasis on retail management practices (e.g. Keh et al., 2006).

In terms of commercial software, we have found one example, ShopSim (Savannah Simulations, 2007), which provides a decision support tool for retail and shopping centre management. It evaluates the layout and design of a shopping centre. The software uses an agent-base approach, where behaviour of agents is driven by survey data. It is a good example of the form of output that we would expect our simulator to create. However, our tool will operate on a departmental level rather than on a shop level, and will investigate different kinds of management practices rather than a shopping centre layout. Furthermore, the input data will come from management and staff surveys in addition to customer surveys.

Choosing a Suitable Modelling Technique

When investigating the behaviour of complex systems the choice of an appropriate modelling technique is very important. In order to make the most suitable selection for our project, we reviewed the relevant literature spanning the fields of Economics, Social Science, Psychology, Retail, Marketing, OR, Artificial Intelligence, and Computer Science. Within these fields a wide variety of approaches is used which can be classified into three main categories: analytical approaches, heuristic approaches, and simulation. In many cases we found that combinations of these were used within a single model (Greasley, 2005; Schwaiger & Stahmer, 2003). From these approaches we identified simulation as best suiting our needs.

Simulation introduces the possibility of a new way of thinking about social and economic processes, based on ideas about the emergence of complex behaviour from relatively simple activities (Simon, 1996). Simulation allows clarification of a theory and investigation of its implications. While analytical models typically aim to explain correlations between variables measured at one single point in time, simulation models are concerned with the development of a system over time. Furthermore, analytical models usually work on a much higher level of abstraction than simulation models.

It is critical to define a simulation model using an appropriate level of abstraction. Csik (2003) states that on the one hand the number of free parameters should be kept as low as possible. On the other hand, too much abstraction and simplification might threaten the fit between reality and the scope of the simulation model. There are several different approaches to simulation, amongst them Discrete Event Simulation (DES), System Dynamics (SD), and ABS. The choice of the most suitable approach always depends on the issues to be investigated, the input data available, the level of analysis and the type of answers that are sought.

Agent-Based Simulation - Our Choice

Although computer simulation has been used widely since the 1960s, ABS only became popular in the early 1990s (Epstein and Axtell 1996). It is described by Jeffrey (2007) as a mindset as much as a technology: "It is the perfect way to view things and understand them by the behaviour of their smallest components". ABS can be used to study how micro-level processes affect macro level outcomes. A complex system is represented by a collection of agents that are programmed to follow simple behavioural rules. Agents can interact with each other and with their environment to produce complex collective behavioural patterns. Macro behaviour is not explicitly simulated; it emerges from the micro-decisions of individual agents (Pourdehnad et al., 2002). Agents have a number of core characteristics: autonomy, the ability to respond flexibly to their environment, and pro-activeness depending on internal and external motivations. They are designed to mimic the attributes and behaviours of their real-world counterparts. The simulation output may be potentially used for explanatory, exploratory and predictive purposes (Twomey & Cadman, 2002). This approach offers a new opportunity to realistically and validly model organisational characters and their interactions, to allow a meaningful investigation of management practices. ABS remains a relatively new simulation technology and its principal application so far has been in academic research. With the availability of more sophisticated modelling tools, things are starting to change (Luck et al., 2005), not to forget the ever-increasing number of computer games using the agent-base approach.

Due to the characteristics of the agents, this modelling approach appears to be more suitable than DES for modelling human-oriented systems. ABS seems to promote a natural form of modelling, as active entities in the live environment are interpreted as actors in the model. There is a structural correspondence between the real system and the model representation, which makes them more intuitive and easier to understand than for example a system of differential equations as used in SD. Hood (1998) emphasised one of the key strengths of ABS, stating that the system as a whole is not constrained to exhibit any particular behaviour, because the system properties emerge from its constituent agent interactions. Consequently assumptions of linearity, equilibrium and so on, are not needed. With regard to disadvantages, there is consensus in the literature that it is difficult to evaluate agent-based models, because the behaviour of the system emerges from the interactions between the individual entities. Furthermore, problems often occur through the lack of adequate real data. A final point to mention is the danger that people new to ABS may expect too much from the models, particularly with regard to predictive ability.

SIMULATION MODEL DESIGN AND IMPLEMENTATION

Modelling Concepts

Case studies were undertaken in 4 departments across 2 retail branches of the same company. An integrative approach to data collection and analysis resulted in a huge set of data and insights. Data collection techniques included informal participant observations, interviews, questionnaires, and we also identified and collated relevant company documentation. Feedback reports and presentations were given to employees with extensive experience and knowledge of the 4 departments to validate our understanding and conclusions. This approach has enabled us to acquire a valid and reliable understanding of how the real system operates, revealing insights into the working of the system as well as the behaviour of and interactions between the different agents within it. We have designed the simulation by applying a DES approach to conceptualise and model the system, and then an agent-based approach to conceptualise and model the actors within it. This method made it easier to design the model, and is possible because only the actors' action requires an agent-base approach.

In terms of performance indicators, these are identical to those of a DES model, however, ABS can offer further insights beyond this. A simulation model can detect unintended consequences, which have been referred to as 'emergent behaviour' (Gilbert & Troitzsch, 2005). Such unintended consequences can be difficult to understand because they are not defined in the same way as the system inputs; however it is critical to fully understand all system outputs to be able to accurately draw comparisons between the relative efficiencies of competing systems. Individuals with extensive experience and knowledge of the real system are essential to figuring out the practical meaning of unidentified system outputs.

Our initial ideas for the simulator are shown in Figure 1. Within our initial simulation model we have three different types of agents (customers, sales staff, and managers), each with a different set of relevant parameters. We will use probabilities and frequency distributions to assign slightly different values to each individual agent. In this way a population is created that reflects the variations in attitudes and behaviours of their real human counterparts. We will need global parameters which can influence any aspect of the system, and we will also define the number of agents in it. Regarding system outputs, we aim to find some emergent behaviour on a macro level. Visual representation of the simulated system and its actors will allow us to monitor and better understand the interactions of entities within the system. Coupled with the standard DES performance measures, we aim to identify bottlenecks to assist with optimisation of the modelled system.

Fig. 1. Conceptual model of our simulator.

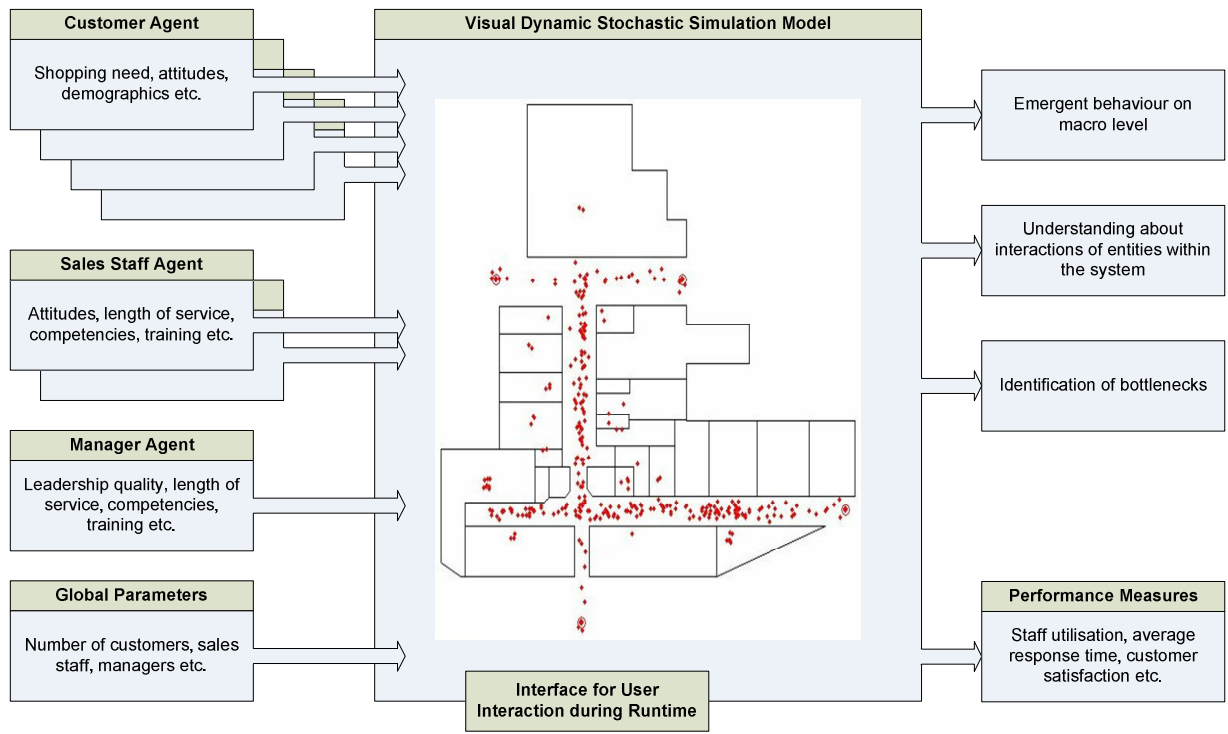

Agent Design

For the conceptual design of our agents we have used state charts. State charts show the different states an entity can be in, and define the events that cause a transition from one state to another. This is exactly the information we need in order to represent our agents later within the simulation environment. Furthermore, this form of graphical representation is also helpful for validating the agent design because it can be easily understood by non-specialists.

The art of modelling pivots on simplification and abstraction (Shannon, 1975). A model is always a restricted version of the real world, and computer modellers have to identify the most important components of a system to build effective models. In our case, instead of looking for components, we have to identify the most important behaviours of an actor and the triggers that initiate a move from one state to another; for example when a certain period of time has elapsed, or at the request of another agent. We have developed state charts for all of the agents in our model: customers, selling staff, and managers. Figure 2 presents one of the state charts, in this case for a customer agent (transition rules have been omitted for simplicity).

Fig. 2. Customer agent conceptual model

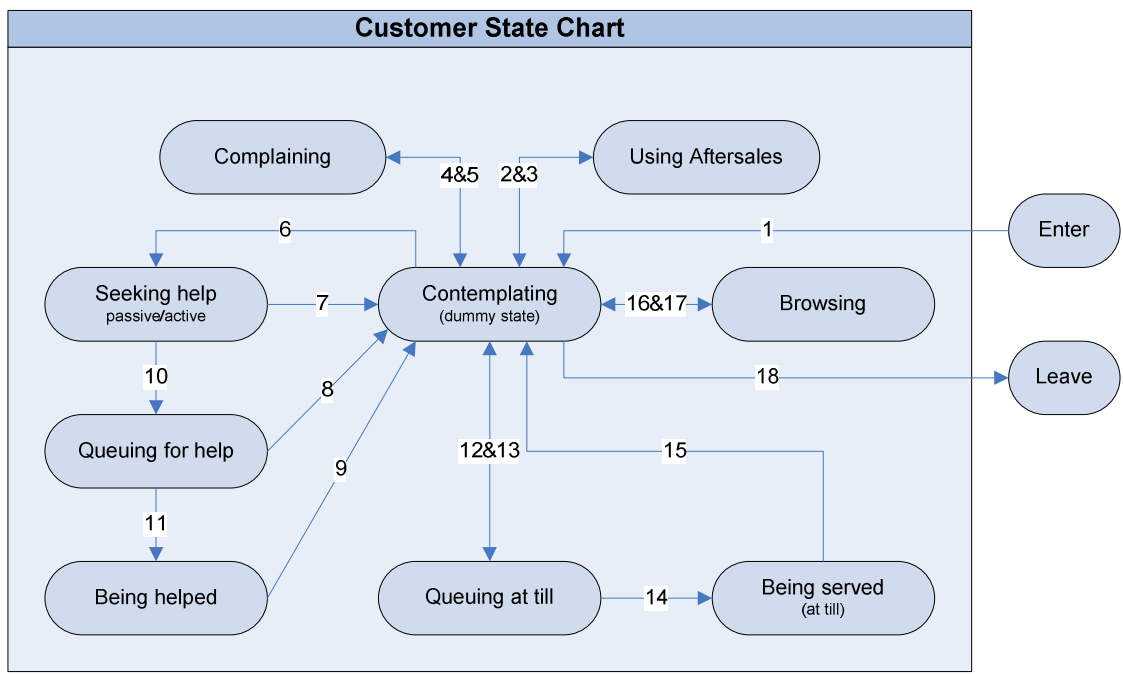

When a customer enters the department she or he is in the 'contemplating' state. This is a dummy state which simply links most of the other states and allows state transition in a straightforward manner. She or he will probably start browsing and after a certain amount of time (delay value derived from a probability distribution) she or he may require help, queue at the till or leave the shop. If the customer requires help she or he will consider what to do and seeks help by sending a signal to a staff member and waiting for a reply. If no staff member is available she or he has to wait (queue) for help, otherwise she or he will spend no time in this state. Subsequently, the customer receives help. Whilst waiting, she or he may browse for another item, proceed to the till to buy a chosen item, or may leave the shop prematurely if the wait is too long.

Implementation

Our simulation has been implemented in AnyLogic™ which is a Java™ based multi-paradigm simulation software (XJ Technologies, 2007). Throughout the model's implementation we have used the knowledge, experience and data gained through our applied case study work. Currently the simulator represents action by the following actors: customers, service staff (including cashiers and selling staff of two different training levels) and section managers. Often agents are based on analytical models or heuristics and in the absences of adequate empirical data theoretical models are employed. However, for our agents we use frequency distributions to represent state change delays and probability distributions for decision making processes. These statistical distributions are the most suitable format to represent the data we have gathered during our case study due to their numerical nature. In this way a population is created that reflects the variations in attitudes and behaviours of their real human counterparts.

Figure 3 presents a screenshot of the current customer agent and staff agent logic as it has been implemented in AnyLogic™. Boxes show customer and staff states, arrows the possible transitions and numbers reveal satisfaction weights.

Fig. 3. Customer (left) and staff (right) agent logic implementation in AnyLogic™

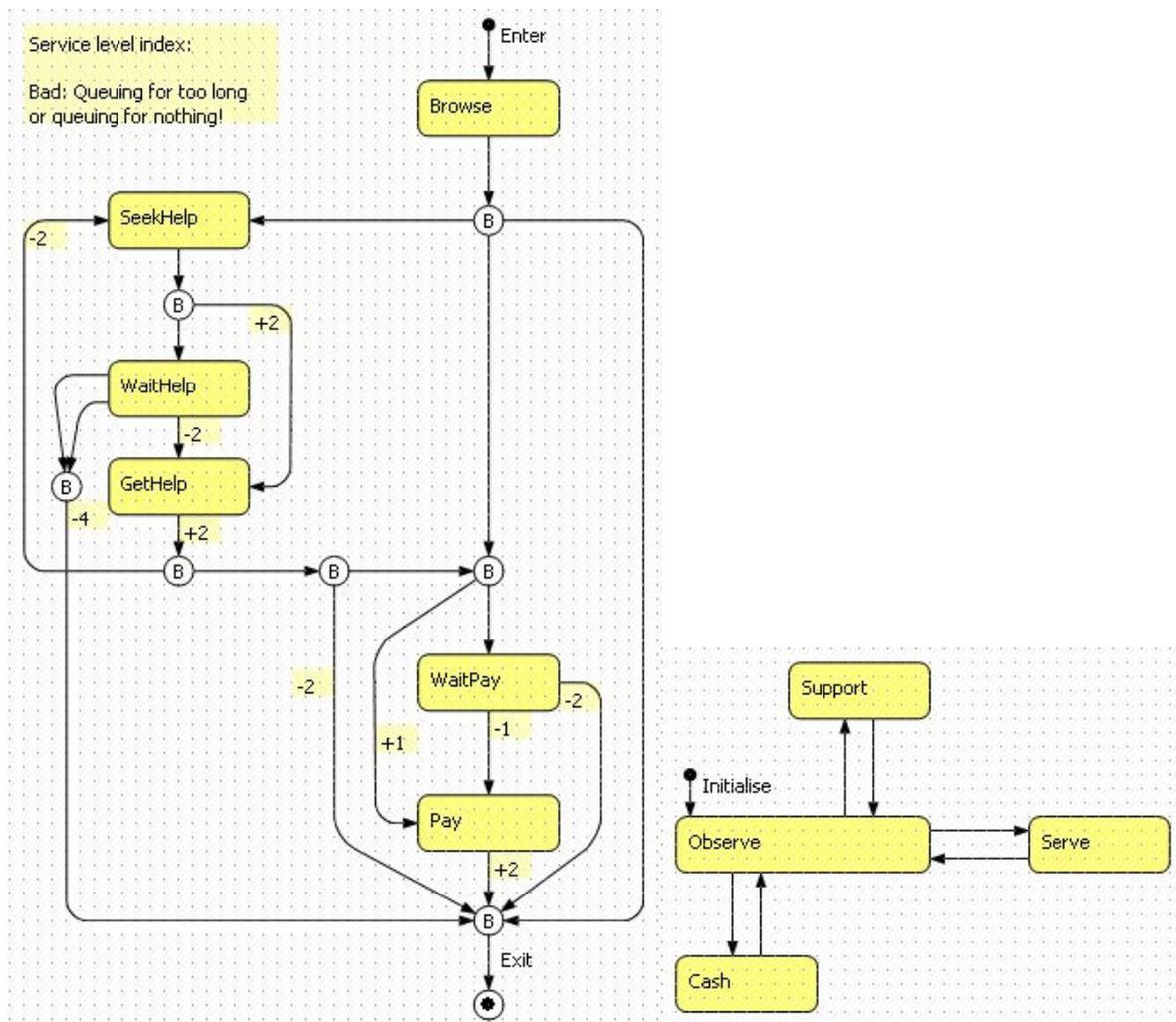

The customer agent template consists of two main blocks which use a very similar logic. In each block, in the first instance, a customer will try to obtain service directly and if he or she cannot obtain it (no suitable staff member is available) he or she will have to queue. The customer will then either be served as soon as an appropriate staff member becomes available, or they will leave the queue if they do not want to wait any longer (an autonomous decision). A complex queuing system has been implemented to support different queuing rules.

The staff agent template, in comparison to the customer agent template, is relatively simple. Whenever a customer requests a service and the staff member is available and has the right level of expertise for the task requested, the staff member commences this activity until the customer releases the staff member.

We introduce a service level index as a novel performance measure using the satisfaction weights mentioned earlier. This index allows customer service satisfaction to be recorded throughout the simulated lifetime. The idea is that certain situations exert a bigger impact on customer satisfaction than others, and we can assign weights to events to account for this. This helps the analyst to find out to what extent customers had a positive or negative shopping experience. It also allows the analyst to put emphasis on different operational aspects of the system, and try out the impact of different strategies.

TEST OF THE SIMULATOR

In order to test the operation of our simulator and ascertain face validity we have completed several experiments. Each experiment is grounded in both theory and practice. We have built the simulation model to allow us to observe and evaluate the impact of different training practices in terms of multiple outcome variables, specifically the volume of sales transactions and various customer satisfaction indices. We use different setup parameters for the departments to reflect real differences between them, for example different customer arrival rates and different service times. All experiments hold the overall number of staffing resources constant at 10 staff and we run our simulation for a period of 10 weeks.

In the first experiment we varied the number of tills open and consequently a different number of selling staff on the shop floor. Hypotheses predicting a curvilinear relationship between the experimental variable and outcome variables were largely confirmed. We expected this type of relationship because the number of normal staff available to provide customer advice and to process purchase transactions must eventually reduce to the extent where there will be a detrimental impact on customer transaction volume and customer perceptions of satisfaction. Further to this, we predicted that the peak level of performance outcomes would occur with a smaller number of cashiers in Audio & TV (A&TV) department as compared to WomensWear (WW) department. This argument is based again on the greater customer service requirement in A&TV, and the higher frequency of sales transactions in WW. Our results supported this hypothesis for both customer satisfaction indices, but surprisingly not for the number of sales transactions where the peak level was at the same point. This is unexpected because we would have anticipated the longer average service times in A&TV to put a greater 'squeeze' on customer advice that is required before most purchases. We would have expected an effect with even a relatively small increase in the number of cashiers.

Our second set of experiments investigated the impact the availability of expert staff on customer satisfaction, predicting a positive and linear relationship (NB this model did not incorporate costs). Results provided some limited support for our hypothesis; however it is likely that the differences reported were too subtle to reach statistical significance. Sales staff working at the case study organisation reported low probabilities for the occurrence of normal staff member support requests (10% in A&TV and 5% in WW), and it is this reality which appears to account for the nominal impact of the number of experts on overall customer satisfaction.

CONCLUSIONS AND FURTHER RESEARCH

In this chapter we have presented the conceptual design, implementation and operation of a retail branch simulator used to understand the impact of management practices on retail performance. As far as we are aware this is the first time researchers have applied an agent-based approach to simulate management practices such as training and empowerment. Although our simulator draws upon specific case studies as source of information, we believe that the general model could be adapted to other retail companies and other areas of management practices that involve a lot of human interaction.

Currently we are building more complex operational features into the simulation model to make it a more realistic representation closer to the real retail environment that we have observed, and we are developing our agents with the goal of enhancing their intelligence and heterogeneity. To meet these goals we are introducing the empowerment of staff, schedules, customer stereotypes and system evolution.

Overall, we believe that organisational psychologists and others should become involved in this approach to understanding behaviour in organisations. In our view, the main potential benefits from adopting this approach must be that it improves our understanding of, and debate about, a problem domain. The nature of the method forces researchers to be explicit about the rules underlying behaviour. Applying ABS opens up new choices and opportunities for ways of working. These opportunities may otherwise be left unnoticed.

XJ Technologies (2007) XJ Technologies - simulation software and services. Available via

<http://www.xjtek.com>. [Accessed Feb 5, 2007].

TERMS AND DEFINITIONS

==Agent:== The word agent has many meanings in Operational Research. In our work, an agent refers to a single entity that will be simulated, i.e. one sales staff, one manager or one customer. Agents are modelled through state charts, describing how, when and why they can change their behaviour.

==Multi-Agent Model:== To simulate a real-life situation a single agent is not sufficient; populations of agents are needed to model most scenarios. For instance, we need a population of customers, a group of managers and various different sales staff. What makes multi-agent models interesting is that the agents do not exist in isolation or a linear fashion, but are able to communicate with and respond to other agents.

==Agent-Based Simulation:== A relatively recent addition to the set of decision support tools. Agent-based simulation extends earlier simulation techniques by allowing the entities that are simulated to make 'autonomous decisions' and to 'negotiate' with other agents. For example, sales staff may decide when to take break times depending on how busy the shop is.

==Management practices:== This usually refers to the working methods and innovations that managers use to improve the effectiveness of work systems. Common management practices include: empowering staff, training staff, introducing schemes for improving quality, and introducing various forms of new technology.